\pdfoutput=1

\documentclass[11pt]{article}

\usepackage{naacl2021}
\usepackage{graphicx}
\usepackage{footmisc}

\usepackage{times}
\usepackage{latexsym}

\usepackage[T1]{fontenc}

\usepackage[utf8]{inputenc}

\usepackage{microtype}

\title{RUBERT: A Bilingual Roman Urdu BERT Using Cross Lingual Transfer Learning}

\author{Usama Khalid \\
  AIM Lab, NUCES (FAST)\\
  Islamabad, Pakistan \\
  \texttt{usama.khalid@nu.edu.pk} \\\And
Mirza Omer Beg \\
  AIM Lab, NUCES (FAST)\\
  Islamabad, Pakistan \\
  \texttt{omer.beg@nu.edu.pk} \\\AND
Muhammad Umair Arshad\\
  AIM Lab, NUCES (FAST)\\
  Islamabad, Pakistan \\
  \texttt{umair.arshad@nu.edu.pk}\\}
  
\begin{document}
\maketitle
\begin{abstract}
In recent studies it has been shown that Multilingual language models under perform their monolingual counterparts \citep{conneau2019unsupervised}. It is also a well known fact that training and maintaining monolingual models for each language is a costly and time consuming process. Roman Urdu is a \textit{resource starved language} used popularly on social media platforms and chat apps. In this research we propose a novel dataset of scraped tweets containing 54M tokens and 3M sentences. Additionally we also propose RUBERT a bilingual Roman Urdu model created by additional pretraining of English BERT \citep{devlin2019bert}. We compare its performance with a monolingual Roman Urdu BERT trained from scratch and a multilingual Roman Urdu BERT created by additional pretraining of Multilingual BERT (mBERT \citep{devlin2019bert}). We show through our experiments that additional pretraining of the English BERT produces the most notable performance improvement. 

\end{abstract}

\begin{figure}[htbp]
\centerline{\includegraphics[scale=.4]{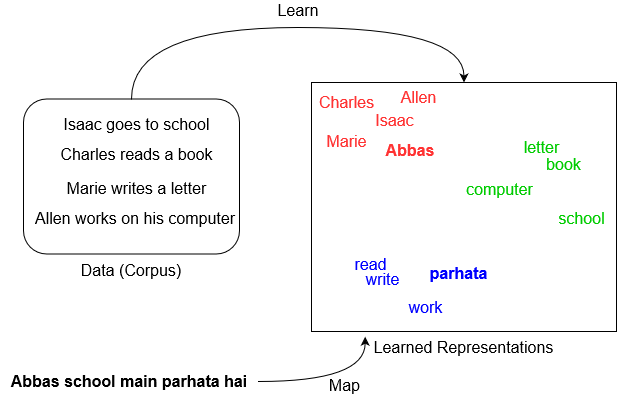}}
\caption{An abstract representation of a cross lingual transfer of a Roman Urdu sentence into the existing space of English learned representations. Note how the existing words are clustered as Nouns(red), Verb(blue) and objects(green) and the words in the Roman Urdu sentence get mapped to those existing spaces.}
\label{fig:overview}
\end{figure}

\section{Introduction}

Roman Urdu is a widely popular language used on many social media platforms and chat apps in South Asian countries like Pakistan and India. However it is a resource starved language, this means that there is not a single corpus, tools or techniques to create Large Pretrained Language models and enable out-of-the-box NLP tasks on Roman Urdu.

\begin{table*}
\centering
\begin{tabular}{llll}
\hline
\textbf{Corpus} & \textbf{Reference} & \textbf{Sentence Count} & \textbf{Word Count}\\
\hline
Urdu NER & \citep{khananamed} & 1,738 & 49,021 \\
COUNTER & \citep{sharjeel2017counter} & 3,587 & 105,124 \\
Urdu Fake News & \citep{amjad2020bend} & 5,722	 & 312,427 \\
Urdu IMDB Reviews & \citep{azam2020sentiment} & 608,693 & 14,474,912 \\
Roman Urdu sentences & \citep{sharf2018performing} & 20,040 & 267,779\\
Roman Urdu Twitter & Proposed & 3,040,153 & 54,622,490\\
\hline
\end{tabular}
\caption{ Statistics of the collected Urdu and Roman Urdu corpora. The Urdu corpora have all been cleaned and transliterated to Roman Urdu. In addition to this a novel corpus for Roman Urdu has also been proposed. }
\label{table:datasets}
\end{table*}

Pretraining a monolingual Language model from scratch requires thousands of GBs of data and tens or hundreds of GPU or TPU hours \citep{yang2019xlnet,liu2019roberta,farooq2019bigdata, zafar2019constructive, zafar2018deceptive, thaver2016pulmonary}. While multilingual pretraining can generally improve the monolingual performance of a model, it is shown that as the number of languages learnt by a multilingual model increases, the capacity available for each language decreases \citep{conneau2019cross}. This phenomenon is also termed as the \textit{curse of multilinguality}. It states that initially while increasing the number of languages contained in a multilingual model leads to better cross-lingual performance, however there comes a stage when overall performance starts to degrade for both monolingual and cross-lingual tasks. Recently a lot of work has been done on multilingual language modeling and it has been shown that monolingual models tend to outperform their multilingual counterparts in similar settings \citep{martin2019camembert,virtanen2019multilingual,pyysalo2020wikibert,beg2007flecs,
koleilat2006watagent,beg2006performance,baigahmed}. Therefore in this research we take a bilingual language modeling approach for the resource starved language Roman Urdu. 

Neural Language modeling was first proposed by \citep{bengio2003neural,collobert2008unified,khawaja2018domain,beg2008critical,
zafar2019using} who showed that training neural networks for predicting the next word given the current, implicitly learnt useful representations, the technique now popularly known as {\it word embeddings}. These embeddings were a leap forward  in the field of NLP by a considerable margin notably after the introduction of techniques like word2vec \citep{mikolov2013distributed}, GloVe \citep{pennington2014glove}, fastText \citep{joulin2017bag}. 

These early techniques \citep{khawaja2018domain} were mostly context-free and a major shortcoming was that they couldn't handle \textit{Polysemy} (one word many meanings), like the word {\it bank} in {\it river bank} and {\it bank account}. Thus started the search for {\it contextual embeddings}. ELMo \citep{peters2018deep} and ULMFiT \citep{howard2018universal} were the first to achieve substantial improvements using LSTM based language models.

Following the line of work of contextual models \citep{beg2009flecs}, a different approach, GPT  \citep{radford2018improving} was proposed to improve the performance on tasks in the GLUE \citep{wang2018glue} benchmark. The new approach to contextual language modelling was to replace LSTMs entirely with then recently released Transformers \citep{vaswani2017attention}. The hugely popular BERT \citep{devlin2019bert} was based on a similar strategy as GPT containing 12 levels but only used the encoder part of the Transformer and looked at sentences bidirectionally.

\begin{figure*}[htbp]
\centerline{\includegraphics[scale=.25]{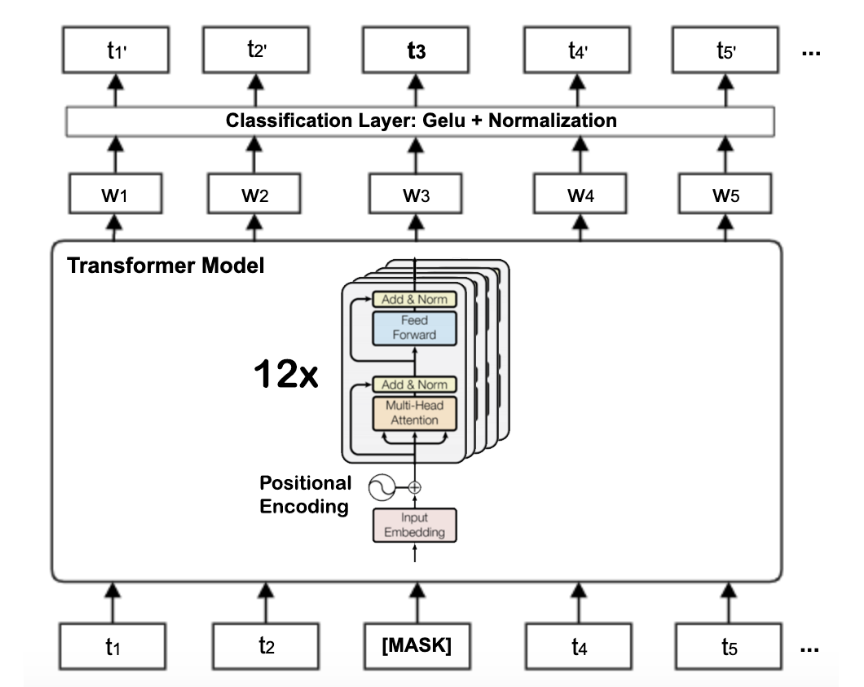}}
\caption{The Transformer based BERT base architecture with twelve encoder blocks.}
\label{fig:architecture}
\end{figure*}

Although large pre-trained Language models \citep{javed2019fairness,beg2001memory} were a huge success, state-of-the-art models were only available for English. A lot of work is being done to reproduce this for other languages but this task is many a times difficult if not impossible for Low resource \citep{seth2006achieving} and resource starved languages like Roman Urdu \citep{zafar2020search}. Multilingual Language modeling aims to solve this by pretraining on a large number of languages \citep{naeem2020deep} to achieve good generalizability \citep{rani2015case} so NLP tasks learned for one language could seamlessly be transferred to multiple languages and also enable zero-shot cross-lingual transfer for new languages. The authors of the original BERT contributed a multilingual version mBERT trained on the Wikipedia of 104 languages \citep{beg2019algorithmic}. This line of work was continued by XLM \citep{conneau2019cross} which achieved state-of-the-art results in cross-lingual classification. This was followed by XLM-R \citep{conneau2019unsupervised} who trained on 2.5TB of text data and outperformed on various multilingual benchmarks \citep{javed2020collaborative}. 

\section{Dataset}
    \label{S-general}

This section explains the data gathering \citep{sahar2019towards} process and the preprocessing steps applied to transform the data in a suitable form to enable the pretraining cycles of BERT and multlingual BERT \citep{qamarrelationship}.

\subsection{Collection}
    \label{S-collection}

Little or no work has been done for Roman Urdu and almost no large publicly available dataset exists for Roman Urdu. Most of our datasets \citep{beg2013constraint} are transliterated from Urdu \citep{nacem2020subspace} using ijunoon's Urdu to Roman Urdu transliteration api \footnote{\url{https://www.ijunoon.com/transliteration/urdu-to-roman/}}. There were also datasets taken from research works on Roman Urdu \citep{sharf2018performing,arshad2019corpus,majeed2020emotion,zahid2020roman}. In addition to the transliterated datasets we propose a novel Roman Urdu dataset consisting of 3 million Roman Urdu sentences and 54 million tokens. The dataset has been scraped from twitter where tweets are partially or wholly written in Roman Urdu. The statistics of the Twitter scraped dataset and all other datsets \citep{uzair2019weec} is shown in Table \ref{table:datasets}. All the above mentioned datasets have also been made publicly available to enable future research for Roman Urdu \footnote{\href{https://drive.google.com/drive/folders/1-7egYe1R9fAanwP2V-UQ3Ca06ZwMkEIf?usp=sharing}{Datasets link}}.

\subsection{Preprocessing}
  \label{S-labels}
  
The datasets from various sources have to be standardized and cleaned to enable them to be passed to pretraining data creation pipeline. The preprocessing pipeline consists of two steps. In the first step data is segmented into sentences using end-of-line character. In the second step all characters except alphabets and numbers are removed, for data transliterated from Urdu this step also includes characters of Urdu script.

\section{Methodology}
      \label{method}

This section explains the steps to transform data for pretraining and defines the architecture and pretraining methods used to enable cross-lingual transfer from English to Roman Urdu and create a Bilingual model from a Monolingual English model.

\subsection{Architecture} 
  \label{S-architecture}
The BERT base architecture consists of 12 layers, 768 hidden nodes, 12 attention heads and 110M parameters \citep{alvi2017ensights}. An overview of this architecture is given in \ref{fig:architecture}. For all our pretraining tasks we use the same BERT base architecture in conjunction with the newly released uncased BERT models \citep{bangash2017methodology}. These Uncased models \citep{karsten2007axiomatic} typically have better performance, but cased versions are useful when case information is useful in a task such as Part-Of-Speech tagging \citep{beg2006maxsm} or Named-Entity-Recognition.

\subsection{Data generation} 

The process of data generation \citep{dilawar2018understanding} for all pretraining cycles \citep{awan2021top} is the same. The data generation process involves two stages. The first stage involves generating a fixed size vocabulary file \citep{tariq2019accurate}, we used the same size file as in English BERT model which is of 30,522 tokens. This was done to enable additional pretraining. In essence the size of the vocabulary of the BERT pretrained model was not changed but the words themselves were. Vocabulary generation or tokenization is performed using the BertWordPieceTokenizer as specified in the official repository \footnote{\label{f1}\url{https://github.com/google-research/bert}}. The tokenizer itself is provided by the HuggingFace team \footnote{\url{https://github.com/huggingface/tokenizers}}.

The second stage in data generation \citep{asad2020deepdetect} involves creating the actual data that will be used in the pretraining task. This requires the vocabulary file as all the tokens from the corpus will be removed except the ones contained in the vocabulary file. For this step we use the \textit{create\_pretraining\_data.py} script provided by github BERT repository \footref{f1}. For larger datasets the pretraining data can be generated in multiple pieces by dividing the original dataset into multiple parts and creating pretraining data for each part.

\subsection{Pretraining} 
  \label{fine-tuning}
The Multilingual BERT \citep{javed2020alphalogger} has currently been trained on 104 Languages \citep{beg2010graph} of which Roman Urdu is not a part of. Thus for the training phase we take a new approach called \textit{additional pretraining} where we run some steps of the pretraining phase for the new language by replacing the old vocabulary entirely by the vocabulary of the new language. Roman Urdu uses the same Latin script as English and many other Languages in addition to this Roman Urdu also consists of many English words such that these are often used interchangeably in a code switched mode. 

This opens up exciting opportunities when building a Roman Urdu Language model from monolingual English BERT. An abstract representation of this idea is also shown in Figure \ref{fig:overview}. Consider the sentence \mbox{\textbf{Abbas school main parhata hai}} (\textit{Abbas teaches in a school}). In this sentence the English BERT knows the context of \textit{school} thus it is able to produce better word representations for surrounding words as compared to the case where not a single word's context is known. 

This notion of cross-lingual transfer can also be applied to Mutilingual models however the learning space is shared with many other languages and \textit{the curse of multilinguality} comes into play. These hypothesis are confirmed by the three types of pretraining Experiments we perform. The first type involves training from scratch. The second type involves additional pretraining of a Monolingual English model and the third type involves additional pretraining of a multilingual model. The results of these experiments are discussed in detail in Section \ref{s-experiment}.
\section{Experiments} 
  \label{s-experiment}
  
In this section we describe the methods used for evaluation and discuss the results obtained for the three different types of pretraining.

\begin{table}
\centering
\begin{tabular}{lll}
\hline
\textbf{Model} & \textbf{MLM} & \textbf{NSP}\\
\hline
RUBERT\textsubscript{Monolingual} & 0.02 & 0.53 \\
RUBERT\textsubscript{Multilingual} & 0.11 & \textbf{0.91} \\
RUBERT\textsubscript{Bilingual} & \textbf{0.23} & \textbf{0.95} \\
BERT\textsubscript{English} & \textbf{0.98} & \textbf{1.0} \\
\hline
\end{tabular}
\caption{ A Comparison between the MLM and NSP accuracies of the three types Roman Urdu model trained and the English BERT model. The Bilingual version of RUBERT achieves the best performance but it is still a long way off from the original English BERT. }
\label{table:results}
\end{table}

\subsection{Evaluation} 

The original BERT model was evaluated mainly on four popular tasks which are GLUE \citep{wang2019glue}, SQuAD  v1.1 \citep{rajpurkar2016squad}, SQuAD v2.0 \citep{rajpurkar2018know} and the SWAG \citep{zellers2018swag} task. However no well know multilingual evaluation task provides support for Roman Urdu. Thus for measuring the performance of our pretrained models we use the evaluation metrics produced in the pretraining phase of each model. The evaluation metrics are based on data that is put aside during creation of the pretraining data.

\subsection{Results} 

The evaluation of all pretraining cycles is performed for two tasks Masked Language Modeling (MLM) and Next Sentence Prediction (NSP) which are the same tasks BERT is trained for. \mbox{Table \ref{table:results}} shows the accuracies of MLM and NSP tasks for the models we trained and also for the Pretrained English BERT model for comparison. The BERT models we have trained using Roman Urdu data are termed as RUBERT. All three pretraining cycles are carried out only using the gathered data mentioned in Section \ref{S-collection}.

The monolingual version of RUBERT is pretrained from scratch while the Multilingual RUBERT is trained by performing additional pretraining steps on a pretrained Multilingual BERT. The Bilingual BERT which achieves the best performance is trained by performing the same additional pretraining steps as before but on a monolingual English BERT Base model. Each of the pretraining and additional pretraining cycles are performed for 10K steps with a learning rate of $2e-5$ except for Monolingual RUBERT which was trained with a learning rate of $1e-4$, the higher learning rate was necessary when training from scratch. 

\section{Conclusion} 
In this research we propose a novel corpus and some transliterated corpora for Roman Urdu a resource starved language. The novel corpus contains sentences extracted from scraped tweets. In addition to this we introduce additional pretraining a new method for cross-lingual transfer of learned representations to a new low resource language. In this research we have shown that Bilingual language modeling is able to outperform multilingual modeling. There are some conditions however to enable cross-lingual transfer which involves the target language to share vocabulary and linguistic features like typography with the source language. We hope that the methods and corpora proposed in this paper will enable further research for resource starved Languages like Roman Urdu.

\bibliography{anthology,custom}
\bibliographystyle{acl_natbib}

\end{document}